\documentclass[sigconf]{acmart}
\AtBeginDocument{%
  }

\newcommand{\keypoint}[1]{\noindent\textbf{#1}\quad}
\newcommand{\cut}[1]{}
\newcommand{\review}[1]{\textcolor{black}{#1}}

\usepackage{amsfonts}
\usepackage{amsmath}
\usepackage{bbold}
\usepackage{subcaption}
\copyrightyear{2026}
\acmYear{2026}
\setcopyright{cc}
\setcctype{by}
\acmConference[SEAMS '26]{21st International Conference on Software Engineering for Adaptive and Self-Managing Systems}{April 13--14, 2026}{Rio de Janeiro, Brazil}
\acmBooktitle{21st International Conference on Software Engineering for Adaptive and Self-Managing Systems (SEAMS '26), April 13--14, 2026, Rio de Janeiro, Brazil}
\acmPrice{}
\acmDOI{10.1145/3788550.3794864}
\acmISBN{979-8-4007-2445-9/2026/04}

\begin{document}

%%
%% The "title" command has an optional parameter,
%% allowing the author to define a "short title" to be used in page headers.
\title{Balancing Multiple Objectives in Urban Traffic Control with Reinforcement Learning from AI Feedback}

%%
%% The "author" command and its associated commands are used to define
%% the authors and their affiliations.
%% Of note is the shared affiliation of the first two authors, and the
%% "authornote" and "authornotemark" commands
%% used to denote shared contribution to the research.
\author{Chenyang Zhao}
% \authornote{Both authors contributed equally to this research.}
\email{zhaoc4@tcd.ie}
% \orcid{1234-5678-9012}
% \author{G.K.M. Tobin}
% \authornotemark[1]
% \email{webmaster@marysville-ohio.com}
\affiliation{%
  \institution{Trinity College Dublin}
  \city{Dublin}
  % \state{Ohio}
  \country{Ireland}
}

\author{Vinny Cahill}
% \authornote{Both authors contributed equally to this research.}
\email{vinny.cahill@tcd.ie}
% \orcid{1234-5678-9012}
% \author{G.K.M. Tobin}
% \authornotemark[1]
% \email{webmaster@marysville-ohio.com}
\affiliation{%
  \institution{Trinity College Dublin}
  \city{Dublin}
  % \state{Ohio}
  \country{Ireland}
}

\author{Ivana Dusparic}
% \authornote{Both authors contributed equally to this research.}
\email{ivana.dusparic@tcd.ie}
% \orcid{1234-5678-9012}
% \author{G.K.M. Tobin}
% \authornotemark[1]
% \email{webmaster@marysville-ohio.com}
\affiliation{%
  \institution{Trinity College Dublin}
  \city{Dublin}
  % \state{Ohio}
  \country{Ireland}
}

%%
%% By default, the full list of authors will be used in the page
%% headers. Often, this list is too long, and will overlap
%% other information printed in the page headers. This command allows
%% the author to define a more concise list
%% of authors' names for this purpose.
% \renewcommand{\shortauthors}{Trovato et al.}

%%
%% The abstract is a short summary of the work to be presented in the
%% article.
\begin{abstract}
Reward design has been one of the central challenges for real world reinforcement learning (RL) deployment, especially in settings with multiple objectives. Preference-based RL offers an appealing alternative by learning from human preferences over pairs of behavioural outcomes. More recently, RL from AI feedback (RLAIF) has demonstrated that large language models (LLMs) can generate preference labels at scale, mitigating the reliance on human annotators. However, existing RLAIF work typically focuses only on single-objective tasks, leaving the open question of how RLAIF handles systems that involve multiple objectives. In such systems trade-offs among conflicting objectives are difficult to specify, and policies risk collapsing into optimizing for a dominant goal. In this paper, we explore the extension of the RLAIF paradigm to multi-objective self-adaptive systems.  We show that multi-objective RLAIF can produce policies that yield balanced trade-offs reflecting different user priorities without laborious reward engineering. We argue that integrating RLAIF into multi-objective RL offers a scalable path toward user-aligned policy learning in domains with inherently conflicting objectives.
\end{abstract}

%%
%% The code below is generated by the tool at http://dl.acm.org/ccs.cfm.
%% Please copy and paste the code instead of the example below.
%%

%%
%% Keywords. The author(s) should pick words that accurately describe
%% the work being presented. Separate the keywords with commas.
% \keywords{Do, Not, Use, This, Code, Put, the, Correct, Terms, for,
  % Your, Paper}
%% A "teaser" image appears between the author and affiliation
%% information and the body of the document, and typically spans the
%% page.

% \received{20 February 2007}
% \received[revised]{12 March 2009}
% \received[accepted]{5 June 2009}

%%
%% This command processes the author and affiliation and title
%% information and builds the first part of the formatted document.
\maketitle

\section{Introduction}
Self-adaptive systems (SAS) increasingly rely on learning-based controllers to operate under uncertainty and evolving requirements. Reinforcement learning (RL), in particular deep RL utilizing deep function approximation, has demonstrated impressive capabilities across a variety of self-adaptive application domains, including traffic signal control \cite{wei2019presslight}, autonomous driving \cite{chan2024safedriverl}, and balloon navigation \cite{bellemare2020autonomous}. However, RL systems rely heavily on a hand-crafted reward functions that encode the designer’s intent as a scalar feedback signal. In practice, specifying a reward function that incentivises the agent to explore the optimal policy while aligning with the system objective is a challenging problem and often requires extensive trial-and-error. Misspecification can induce reward hacking, unsafe behaviours, or brittle policies that overfit to proxy reward signals rather than truly reflect desired behaviour. The challenge becomes more pronounced in multi-objective settings, where performance must be balanced across multiple, potentially conflicting, criteria. Aggregating multiple objectives into a single scalar can be challenging: poorly designed reward functions can lead the learning agent to over-optimize a dominant objective while neglecting others, and may fail to reflect the intent of users. 

To address the challenges in designing reward functions, preference based RL (PbRL) has emerged as a compelling alternative by replacing hand-crafted reward signals with pairwise comparisons of behaviours \cite{christiano2017deep}. Instead of scalar rewards, the agent queries an annotator to indicate which of two presented behaviour trajectories (or segments) is preferred, then learns a reward model that incentivizes the agent to align with those preferences. More recently, RL from AI feedback (RLAIF) leverages large AI systems (generally LLMs) to automate large scale preference data collection. Despite promising results \cite{klissarov2023motif, huang2025trend}, most existing work targets single-objective tasks, where the main challenges are encouraging sufficient exploration or expressing high-level intent as a scalar function. The question of how these approaches extend to multi-objective scenarios, with competing objective priorities, remains largely unexplored.

In this paper, we investigate using LLMs as preference annotators for multi-objective RL (MORL). To address the gap in multi-objective RLAIF, we introduce an annotation workflow specifically tailored to multi-objective scenarios that elicits cross-objective comparisons and conditions labels on user specifications. We empirically show that RLAIF can learn user-aligned policies, without requiring intensive work on reward engineering. \review{We also demonstrate, in a scenario with directly competing objectives, that the policy behaviour can be directed to match desired trade-offs via natural language prompts.} We illustrate the proposed approach in a traffic signal control (TSC) problem, as a canonical example of a multi-objective self-adaptive application. Traffic signal control inherently constitutes a multi-objective optimisation problem, involving competing performance metrics such as traffic throughput and ecological impacts (e.g., emissions). Furthermore, the priorities of different lanes or approaches across the same intersection layout may vary due to road importance or the presence of buses or emergency vehicles. An optimal controller must therefore adapt its behaviour to such variations and deploy different control policies. The inherent trade-offs among these objectives render the problem particularly challenging. Given this complexity, TSC provides an excellent exemplar for applicability of RLAIF algorithms, suitable for testing their performances in managing multiple conflicting objectives under dynamic conditions.

In summary, this paper posits that the recent advancements in LLMs and RLAIF techniques pave a path to a self-managed, verbally-steered learning loop for SAS with multiple objectives. Specifically, stakeholders can express high-level goals and constraints in natural language, and an LLM can be prompted to turn these directives into trade-off-aware preference labels over trajectories. An RL agent can then optimize its behaviour against these generated labels to produce either a single policy aligned with the stakeholder preferences, or a diverse set of policies spanning qualitatively different trade-offs between objectives.

\section{Related Work}
\label{sec:related_work}
In this section, we introduce background work in Preference-based RL (PbRL)  and in utilizing LLMs to provide feedback for RL systems, followed by related work on learning multi-objective self-adaptive systems in traffic signal control.

\subsection{Preference-based Reinforcement Learning}
\label{sec:pbrl}
PbRL has emerged as a principled way to mitigate the challenges of reward engineering in reinforcement learning \cite{wirth2017survey, christiano2017deep, lee2021pebble}. Rather than hand-crafting a reward, PbRL infers a reward model from pairwise preference feedback, typically human comparisons between short trajectory segments or outcomes \cite{christiano2017deep}. This is appealing because humans generally find it more natural to compare behaviours than to specify a precise reward function \cite{wirth2017survey}.
PbRL has recently been extended to multi-objective RL (MORL), where agents must balance multiple criteria. One line of work elicits hierarchical or importance-aware preferences, comparing segments by scanning objectives in a ranked order until a salient difference is found \cite{bukharin2023deep}. Another assumes linearly weighted objectives and trains a weight-conditioned policy by sampling trade-off weights during learning \cite{mu2025preference}. Despite these advances, most prior systems depend on either human annotators or a scripted teacher—i.e., synthetic labels generated from a oracle reward function \cite{lee2021b, lee2021pebble}. To the best of our knowledge, we are the first to explore using LLMs as multi-objective annotators, scaling PbRL to real-world multi-objective problems with far less human effort.

\subsection{Reinforcement Learning From AI Feedback}
\label{sec:rlaif}
The advent of powerful foundation models (e.g., GPTs, Llama) has catalysed efforts to integrate them into RL pipelines to reduce human labour, by directly evaluating trajectories/transitions \cite{kwon2023reward, wu2023read}, helping specify reward functions \cite{ma2023eureka, xie2023text2reward}, or providing action advice \cite{zhou2023large}. These studies show that LLMs can benefit RL without requiring expert demonstrations or extensive reward engineering. Moreover, when explicit reward design is difficult and direct evaluations are noisy, Reinforcement Learning from AI Feedback (RLAIF) leverages such AI systems to generate pairwise preferences at scale, shown success in domains such as NetHack \cite{klissarov2023motif} and robotic manipulation \cite{wang2024rl}. Follow-up studies tackle label quality and robustness via crowd-sourced LLM annotators \cite{wang2025prefclm}, ensembles of reward models \cite{huang2025trend}, and richer supervision such as rating-style annotations \cite{luu2025enhancing}.

However, prior work largely addresses single-objective tasks. In multi-objective settings, the preferred trade-off among objectives can vary across annotators. This heterogeneity induces noisy and inconsistent preference labels, complicating aggregation and reward learning in RLAIF. In this paper, we focus on multi-objective scenarios and study how to obtain and use LLM-generated preferences that faithfully reflect these trade-offs while enabling stable, scalable policy optimisation.

\subsection{Multi Objective Traffic Signal Control}
\label{sec:motsc}
Intelligent traffic signal control (TSC) aims to adaptively adjust signal timings based on real-time traffic demand \cite{zhao2024survey}. Deep RL approaches are frequently used in  TSC systems due to their adaptability \cite{wei2021recent}. Different approaches use different reward designs to achieve the expected traffic behaviours, e.g., vehicle waiting time \cite{gao2017adaptive}, queue length \cite{zheng2019diagnosing}, and junction pressure \cite{wei2019presslight}. Recently, growing sustainability concerns have driven growing interest in multi-objective TSC which jointly optimises traffic efficiency, environmental impact, and fairness across stakeholders. Most methods use scalarization, typically combining different objectives in a weighted-sum reward, e.g., minimizing delay while penalizing carbon emission consumption \cite{zhang2024multi, zhang2024mmd}. However, such approaches often require extensive weight tuning to find the balance between the objectives. Other work explores Pareto-based techniques that learn sets of policies spanning different trade-offs\cite{saiki2023flexible}, which provides flexibility in policy choices, at the cost of high computational resources. In this work, we explore the use of PbRL and AI feedback in addressing the multi-objective problem in TSC, aiming to reduce weight-tuning effort without incurring the computational burden of training an entire set of policies.

\begin{figure*}[t]
    \centering
    \includegraphics[width=0.65\linewidth]{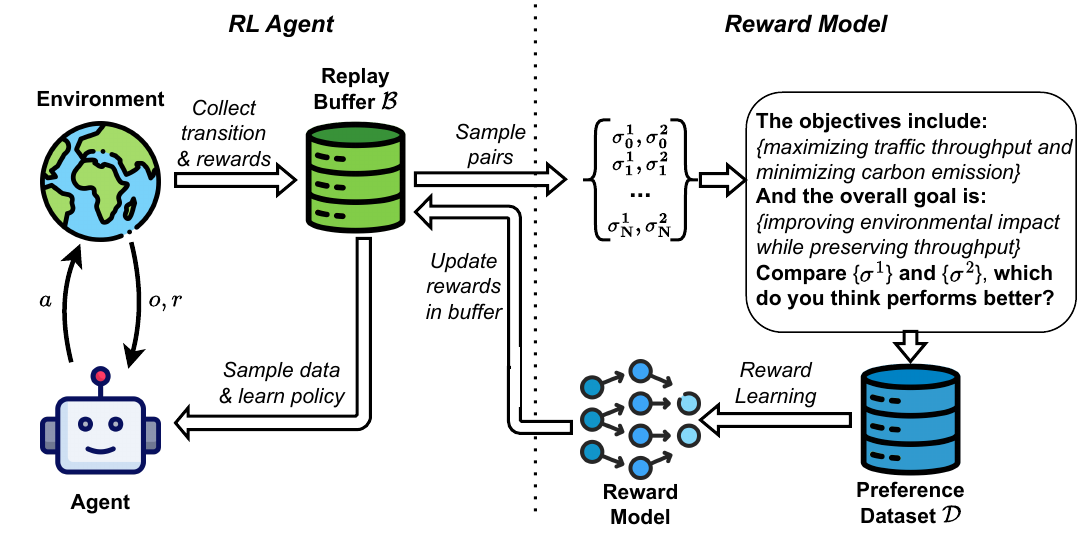}
    \vspace{-0.4cm}
    \caption{Our framework of RLAIF for multi-objective tasks: the policy interacts with the environment, transitions enter the replay buffer $\mathcal B$; segment pairs are annotated by an LLM and stored into a preference buffer $\mathcal D$; the reward model $r_\psi$ updates from $\mathcal D$; re-scores $\mathcal B$, and the policy is trained with RL algorithms with the updated replay buffer. During annotation, we make the optimization criteria explicit, stating the task objectives and desired trade-offs, so the learned reward reflects user expectations and aligns behaviour across multiple objectives without bespoke reward engineering.}
    \label{fig:framework}
    \vspace{-0.3cm}
\end{figure*}

\section{LLMs for RLAIF in Multi-objective Tasks}
\label{sec:rlafi_mo}
In this section, we present our proposed RLAIF for multi-objective tasks framework. We tailor the annotation prompt template, explicitly stating the task objectives and desired trade-offs via natural language, such that annotations better reflects user expectation over multiple objectives.

Prior work \cite{wang2024rl, klissarov2023motif} targets a single objective, e.g., opening a drawer with a robot arm or advancing to the next level in a game. Annotators can judge “how well the task was achieved” along one scalar criterion. In multi-objective settings, especially when objectives conflict, there is usually no single globally preferred solution. Consequently, annotator preferences become conditional on implicit weights over objectives, which vary across annotators and over time, yielding more ties and noisier, inconsistent labels. This in turn can destabilize reward-model training. To mitigate this issue, we propose the use of more structured prompts for LLM annotators tailored for multi-objective tasks, instructing them to evaluate the performances along each of the objectives and the overall alignment with user's expectation.

Fig.~\ref{fig:framework} provides an overview of our multi-objective RLAIF framework. The system alternates between data collection, preference-based reward learning, and policy learning. Provided with proper prompts about different task objectives and the overall goal, LLMs can serve as annotators to autonomously generate preference labels. \review{Specifically, the current policy interacts with the environment to produce transitions $(s_t, a_t, s_{t+1})$, which are stored in an ordered replay buffer $\mathcal{B}$. Periodically, pairs of trajectory segments, i.e., a sequence of transitions, are sampled from $\mathcal{B}$ and an LLM queried with an objective prompt to obtain preference labels; uncertain judgments are filtered and the remaining labels are stored in a preference buffer $\mathcal{D}$}. A reward model $r_\psi$ is then trained on these pairwise labels. After each update of the reward model, the transitions in $\mathcal{B}$ are relabelled with new rewards using $r_\psi$. The policy $\pi_\theta$ is updated with standard off-policy RL algorithms (DQN in our implementation) using rewards provided by $r_\psi$. 

\subsection{Preference Annotation}

We modify the prompting template used in previous work \cite{wang2024rl} for generating preference labels via LLMs, so as to explicitly address the multi-objective nature of tasks and the users' expectations. Specifically, the multi-stage chain-of-thought annotation process comprises the following steps:

\keypoint{Observation analysis} Given a verbal instruction that specifies the target objectives, the LLM is prompted to examine a pair of observation segments $\sigma^1$ and $\sigma^2$ independently. To address the multiple objective nature, the LLM is prompted to evaluate each segment's performance with respect to each individual objective, as well as how well the segment performs in balancing across objectives and addressing the overall goal.  

\keypoint{Segment comparison} Next, the LLM compares the two segments $\sigma^1$ and $\sigma^2$, drawing on both the raw observation data and the per-segment analyses generated in the previous step, to judge which segment better satisfies the overall goal instructions. 

\keypoint{Preference labelling} Finally, the LLM is prompted to assign a preference label $y\in\{0,1,2\}$, where $1$ indicates a clear preference for the first segment \( \sigma^1\), $2$ indicates a clear preference for the second segment \( \sigma^2\), and $0$ indicates that the LLM is not confident in assigning a preference. Pairs labelled with $0$ are excluded from the preference dataset, to improve the robustness of reward learning.  

\begin{figure}[t]
\centering
\includegraphics[width=\linewidth]{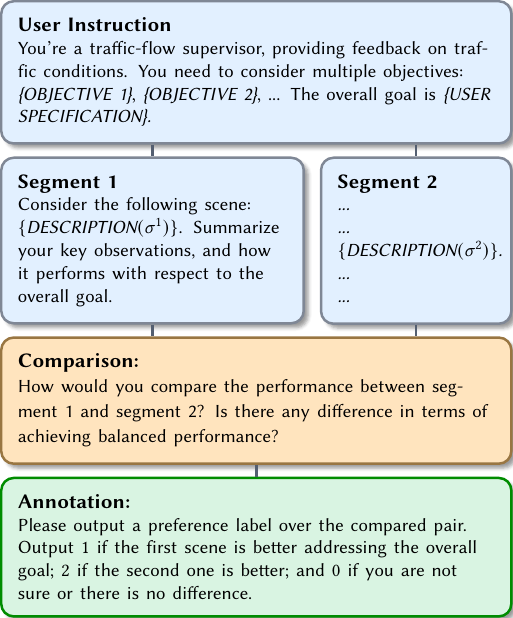}
\vspace{-0.3cm}
\caption{An illustrated workflow of the preference annotation process. The annotator is given considered objectives, the overall goal, and descriptions of the sampled pair of segments $\{\sigma^1, \sigma^2\}$, and outputs the preference label $y\in\{0,1,2\}$.}
\label{fig:pref_annotation}
\vspace{-0.3cm}
\end{figure}

The full annotation process is illustrated in Figure~\ref{fig:pref_annotation}. When querying for annotation, a pair of segments is sampled, the chain-of-thought annotation workflow above applied, and the resulting label stored into the preference buffer \( \mathcal D \). Note that the human only participates before the start of the training process, providing instructions on task objectives. Thereafter, all prompts are generated by the proposed framework and annotations during training are generated autonomously by the LLM.

\subsection{Reward and Policy Learning}

\review{Given collected transitions and preferences, the reward model is trained to approximate the underlying LLM preference distribution and the policy model is trained to maximize accumulated return. Following Bradley-Terry model \cite{bradley1952rank}, the estimated preference distribution over a pair $\{\sigma_1, \sigma_2\}$ given a reward model $r_\psi$ is described as  
$P_\psi(\sigma^1 \succ \sigma^2)
 = \text{softmax}\!\left(\left(r_\psi(\sigma^1) - r_\psi(\sigma^2)\right)/\beta\right)$. The objective of reward learning is to minimise the cross-entropy loss between estimated preference distribution and true labels:}
\begin{equation*}
\max_{\psi} \mathbb{E}_D\Big[
\mathbb{1}\{y_i=1\}\,\log P_\psi(\sigma_i^1 \succ \sigma_i^2)
+\mathbb{1}\{y_i=2\}\,\log P_\psi(\sigma_i^2 \succ \sigma_i^1)
\Big].
\end{equation*}
For policy learning, DQN serves as the underlying RL algorithm. To bootstrap, we first collect a small replay buffer by training the agent under an arbitrary proxy reward and query preferences on segment pairs from this buffer to fit an initial reward model. During training, comparison pairs are continually sampled from the evolving replay buffer. To address non-stationarity in rewards, we relabel the rewards of all stored transitions in the replay buffer with the latest reward model whenever it is updated, to ensure that rewards always are specified with the same reward model across the entire replay buffer \cite{lee2021pebble, wang2024rl}.

% \definecolor{navy}{RGB}{41,55,78}
% \definecolor{sky}{RGB}{226,240,255}
% \definecolor{amber}{RGB}{255,196,110}
% \definecolor{mint}{RGB}{217,244,226}

% \tikzset{
%   >={Triangle[length=6pt,width=8pt]},
%   font=\sffamily,
%   box/.style={
%     draw=navy!60, very thick, rounded corners=6pt,
%     fill=sky, text=black, align=left, inner sep=8pt,
%     text width=0.85\linewidth, blur shadow
%   },
%   smbox1/.style={
%     box, text width=0.5\linewidth
%   },
%   smbox2/.style={
%     box, text width=0.25\linewidth
%   },
%   compare/.style={
%     box, fill=amber!45, draw=amber!60!black
%   },
%   final/.style={
%     box, fill=mint, draw=green!55!black
%   },
%   arrow/.style={
%     navy!80, very thick
%   }
% }

\section{RLAIF for Multi-objective TSC}
\label{sec:exp}

To demonstrate the effectiveness of RLAIF in a self-managed system with multiple objectives, we test the framework in traffic signal control (TSC) scenarios \footnote{Our code is publicly available on \url{https://github.com/cyzhao1991/RLAIF_MO_TSC}.}. TSC regulates phase timing at intersections to manage vehicle flows. The problem is inherently involving multiple objectives, including traffic throughputs along different junction approaches and carbon emissions.

\subsection{Experiment Setup}
\keypoint{Environment formulation} We evaluate RLAIF on a single four-way intersection simulated in Simulation of Urban MObility (SUMO) \cite{SUMO2018}. Each incoming approach comprises two lanes: one dedicated to left-turn movements and another for through and right-turn movements. In the experiments, each episode simulates the traffic for 10,000 seconds and each environment step corresponds to 5 seconds of real-time traffic flow. We follow the same configuration for traffic patterns as \cite{sumorl}, with 200 vehicles per hour on each lane in the north-south (NS) direction and 600 each in the east-west (EW) direction. Signal actuation is framed as a four-phase discrete action space, \{NS left-turn, NS through and right-turn, EW left-turn, EW through and right-turn\}. Each action sets the lights for its direction as green and others as red. Note that a yellow phase of 2 seconds is automatically inserted before a red phase. \review{The observation is defined as [signal phase, min green, each lane vehicle density, each lane queue], where signal phase is a one-hot encoded vector indicating the current active green phase and min green is a binary value indicating whether minimal time for a green phase has already passed. This observation encodes information about both light phases and lane-level traffic. }

%\footnotetext{Code is available at \url{https://github.com/cyzhao1991/RLAIF_MO_TSC}.}

\begin{figure}[t]
  \centering
    \includegraphics[width=\linewidth]{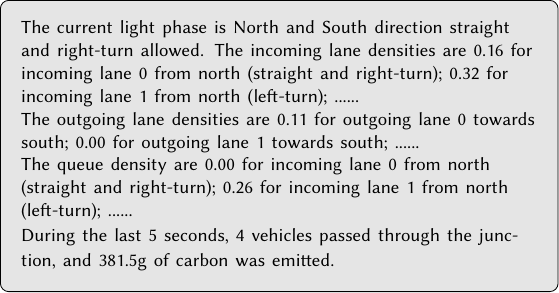}
  % \begin{tikzpicture}
  %   \node[draw=black,
  %         fill=gray!20,        % grey background
  %         rounded corners,      % optional: rounded corners
  %         inner sep=1em,        % space inside box
  %         text width=0.92\linewidth, % width of textbox
  %         align=left]           % alignment of text
  %        (greybox)
  %        {\small
  %          The current light phase is North and South direction straight and right-turn allowed. The incoming lane densities are 0.16 for incoming lane 0 from north (straight and right-turn); 0.32 for incoming lane 1 from north (left-turn); ...... \\
  %          The outgoing lane densities are 0.11 for outgoing lane 0 towards south; 0.00 for outgoing lane 1 towards south; ......\\
  %          The queue density are 0.00 for incoming lane 0 from north (straight and right-turn); 0.26 for incoming lane 1 from north (left-turn); ......\\
  %          During the last 5 seconds, 4 vehicles passed through the junction, and 381.5g of carbon was emitted.
  %        };
  % \end{tikzpicture}
  \vspace{-0.5cm}
  \caption{An exemplar text description of a traffic segment with length 1 in the task of balancing traffic throughput and environmental impact. The text description includes the observations, explained with their semantic meanings, as well as information on throughput and carbon emissions.}
  \label{fig:trans}
  \vspace{-0.5cm}

\end{figure}

\keypoint{Implementation details} To translate simulator observations into natural language, we use a rule-based template that maps each element of the observation vector to a human-readable description and adds auxiliary context, e.g., the number of vehicles that passed in the last time step. An example translation in the task of balancing throughput and environmental impact is included in Figure ~\ref{fig:trans}.

This representation helps LLMs assess traffic conditions and provide consistent, interpretable preferences annotations with brief justifications. As for the prompts, we use a universal prompt template as described in Fig.~\ref{fig:pref_annotation} and gpt-4.1-nano \cite{openai_gpt_4_1_nano_2025} as the underlying language model, with only the \textit{OBJECTIVES}, \textit{USER SPECIFICATION}, and the text description for each compared segment \textit{DESCRIPTION}($\sigma$) varying across queries.

Both the reward model and the Q network (policy model) use a 2-layer MLP architecture with 256 hidden units per layer and LeakyRelu activation. The reward model uses an ensemble of three networks with linear output heads and no activations. We use AdamW \cite{loshchilov2017decoupled} as the optimizer with a weight decay coefficient of $0.01$, a learning rate of $3e-4$, and a batch size of $128$. The frequency at which we collect feedback is 5000 environment steps. We follow the disagreement-based strategy when sampling pairs for comparison, where pairs that models in the ensemble produce the most different estimations are prioritized for annotation. Concretely, at each sampling step, $1024\times5$ pairs are initially sampled from the replay buffer at random. For each sampled pair, three probabilities of $\sigma$ winning are estimated with different reward models in the ensemble using Bradley-Terry model, and the standard deviation over the probabilities is considered as a measure for model disagreement. The sampled batch is then ranked based on their disagreement and the top $1024$ samples are proceeded further for annotation.

\begin{figure*}[t]
    \centering
    \begin{subfigure}{0.33\linewidth}
        \includegraphics[width=\linewidth]{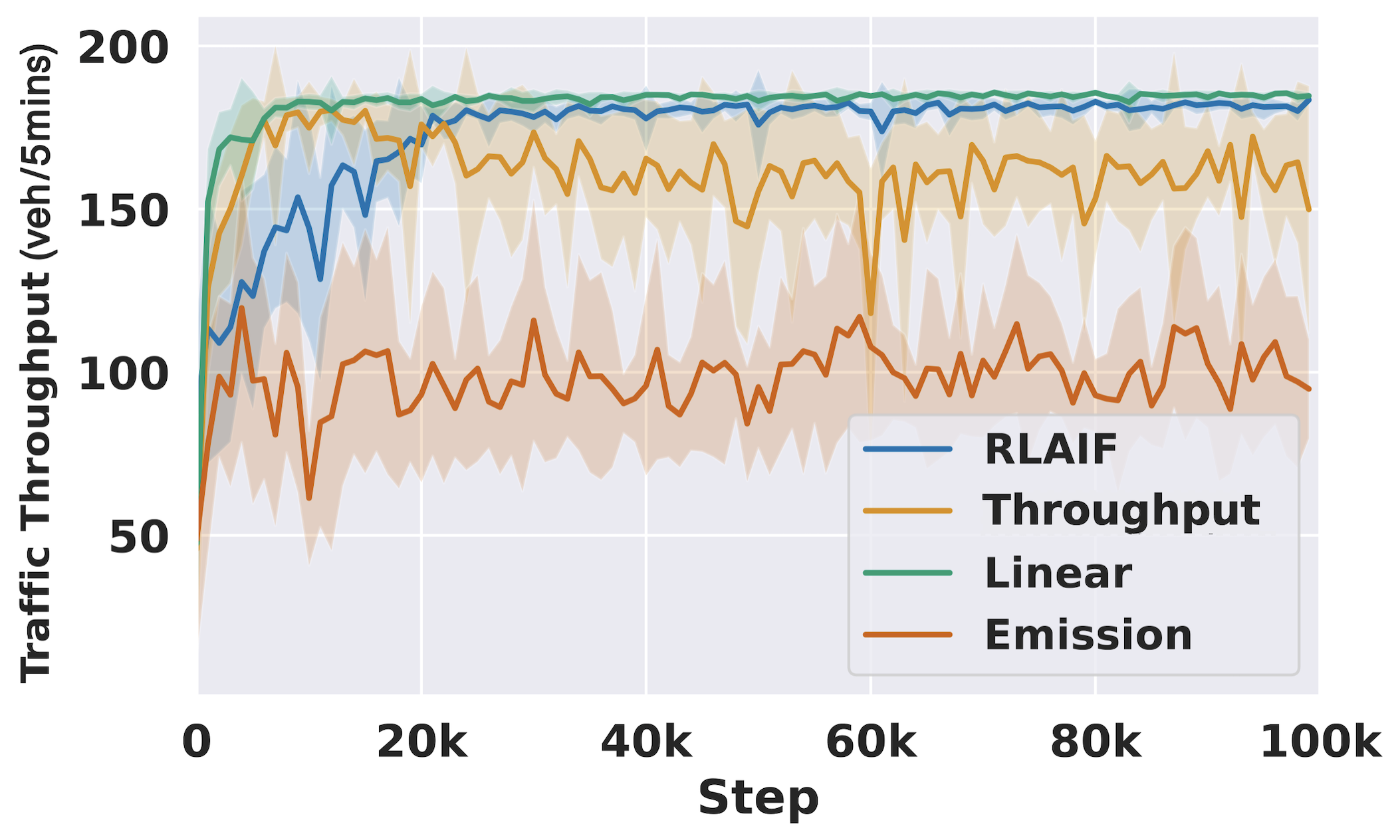}
    \end{subfigure}
    \begin{subfigure}{0.33\linewidth}
        \includegraphics[width=\linewidth]{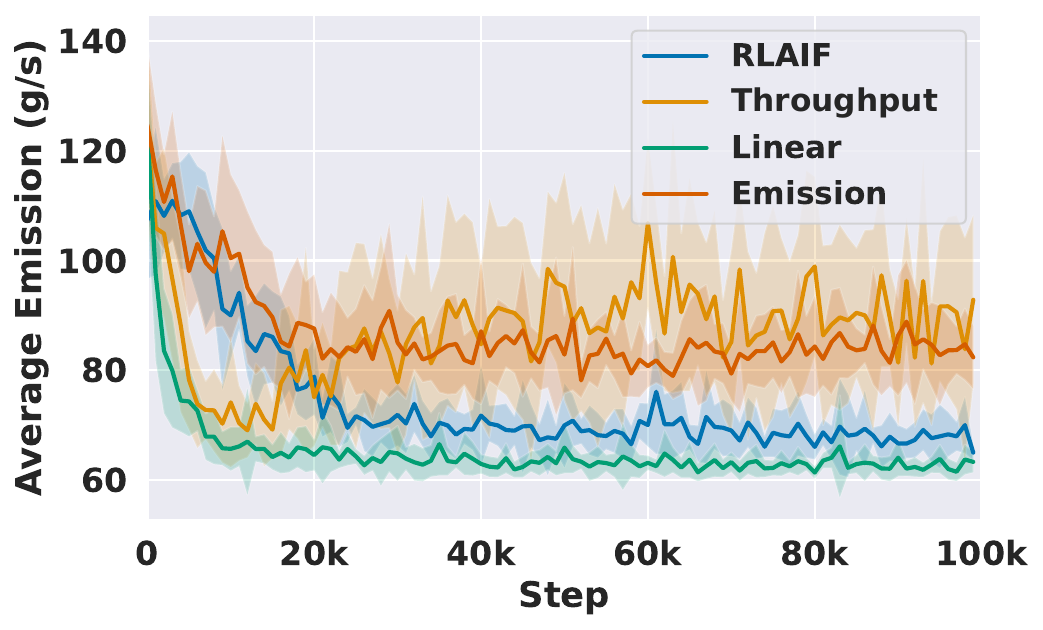}
    \end{subfigure}
    \begin{subfigure}{0.32\linewidth}
        \includegraphics[width=\linewidth]{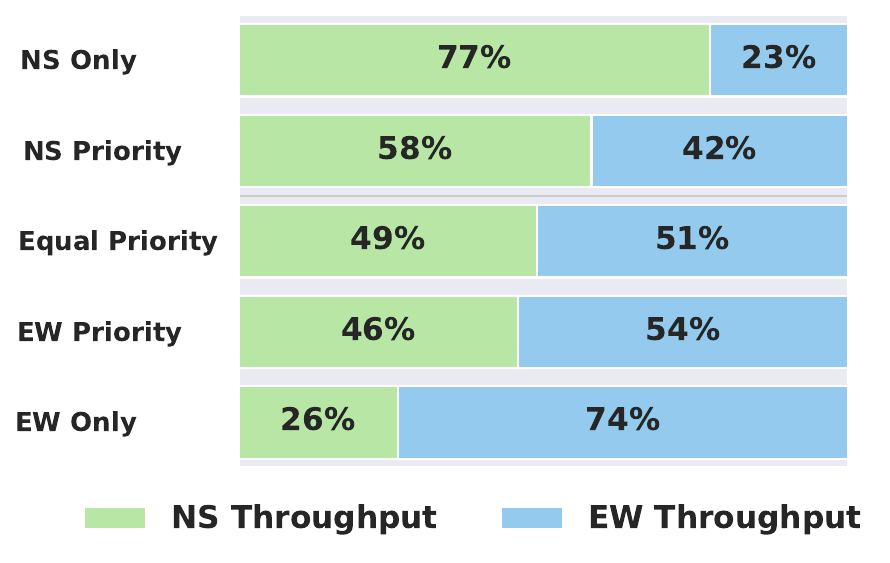}
    \end{subfigure}
    \vspace{-0.4cm}
    \caption{Left and middle: Average traffic throughput and CO2 emission in throughput-emission scenario through learning process. RLAIF learns better performances compared with single objective baselines, and falls short only of the Linear baseline, which requires reward engineering. 
    Right: Comparison between 5 policies trained with different user specifications in lane priorities scenario with instructions towards different lane priorities.}
    \label{fig:carbon}
    \vspace{-0.3cm}
\end{figure*}

\subsection{Experiment Scenarios}
In the experiments, we evaluate RLAIF with the following two scenarios: \textit{throughput-emission} and \textit{lane priorities}.

\keypoint{Throughput-emission} In this setup, we consider two objectives that are most commonly discussed in TSC: \textit{overall throughput}, which is the total number of vehicles exiting the junction, and the total carbon emission, which is calculated based on the HBEFA3 model in SUMO. Specifically, in the prompt template, we specify the OBJECTIVES as \textit{maximizing the throughput} and \textit{minimizing carbon emission} and the USER SPECIFICATION as \textit{improve environmental impact while preserving or only marginally reducing throughput.} We benchmark against a multi-objective baseline with linear reward scalarisation (“Linear”), using a weighted sum of the two pre-defined reward functions $r = \alpha r_1 + (1-\alpha) r_2$, with no learned reward models. $r_1$ is the number of vehicles driving through the junction and $r_2$ the total amount of carbon emissions in grams/second. To find the best weights, we conduct a grid search for $\alpha$, exploring values within the set $\{0.1, 0.2, ...,0.9\}$. \review{We monitor the visualized simulated traffic patterns of different agents as well as their real-time quantitive metrics, including throughput, carbon emission, average vehicle speeds, and queue length, and report the best validation result ($\alpha = 0.7$).} We also include single-objective DQN variants (“Throughput” and “Emission”), which are learned with $r_1$ and $r_2$, respectively.

\keypoint{Lane priorities} 
Furthermore, to assess how prompt specifications influence the agent’s behaviour, we include a scenario with directly competing objectives: throughput along different directions. In this experiment, we deploy a symmetric traffic demand pattern: For all four direction, 400 vehicles go straight or take a right-turn and 300 vehicles take a left-turn every hour. We prompt the LLM annotators with different instructions on lane priorities and evaluate the impact on lane-specific throughput. Concretely, we specify in the prompt template that the OBJECTIVES are \textit{to maximize the throughput along the north-south (east-west) direction}. And the USER SPECIFICATION are \textit{to ensure high throughput along the north-south (east-west) direction since there are possibly emergency vehicles} (NS only and EW only); \textit{to prioritise high throughput along the north-south (east-west) direction since there are buses along the lane} (NS priority and EW priority); and \textit{to ensure balanced throughput along both directions} (Equal priority). 

\vspace{-0.1cm}
\subsection{Results and Discussions}
\keypoint{Learning Performances} Figure~\ref{fig:carbon} shows the  performance in the throughput-emission scenario for RLAIF and three DQN baselines. In early stage of training (<20k steps), RLAIF learns slower compared with other approaches with stationary rewards. We hypothesize that this is because the reward model is evolving with the policy in RLAIF, leading to lower sample efficiency in early stage. However, after convergence (>40k steps), RLAIF achieves better performances against single-objective baselines. 
RLAIF only falls short of the Linear baseline, achieving competitive results in throughput but emitting more carbon by a small margin ($\sim 65(g/s)$ in RLAIF and $\sim 60(g/s)$ in Linear). \review{However, note that Linear baseline is selected via exhaustive grid search over 9 different weight settings, incurring additional simulation steps and human oversight to identify the best scalarization.} \review{With RLAIF, the budget in annotating preferences also need to be considered. In our experiments, about 20k annotations are made each run, which costs around $1-2$ million tokens ($\$0.1 -0.2$ with model gpt-4.1-nano). Out of all LLM annotations, an average of 44\% of them are labelled with $0$ and thus filtered out in the reward model learning step.} Overall, the comparison suggests that RLAIF can converge to a competitive policy without extensive manual reward engineering or human labelling, leveraging AI-based preference signals instead.

\keypoint{Sensitivity to instructions}
The comparisons of RLAIF-learned performances with different lane priorities instructions are included in Figure~\ref{fig:carbon}. When no lane priority is specified, the agent learns almost equal throughput along two directions in \textit{Equal Priority}. When we gradually increase the importance to a specific direction through the prompts, we observe clear changes in the traffic behaviour, with the NS direction throughput moving to $58\%$ when instructed to prioritise north-south direction and $77\%$ to ensure north-south high throughput.
This suggests tuning the prompts as a promising way for steering the expected behaviours of outcome policies in RLAIF, when learning a diverse set of policies is of interest. 
\section{Conclusions}
In this paper, we explore the extension of RLAIF paradigm, which leverages the LLM models for preference annotations in PbRL, to the context of multi-objective self-adaptive systems. We introduce a tailored LLM-annotation workflow for multi-objective scenarios that elicits cross-objective comparisons and conditions labels on user specifications. Our early evaluation in two traffic signal control scenarios shows that RLAIF yields policies with balanced performance across objectives and enables prompt-based tuning of policy behaviour. Overall, we show that recent advances in LLMs and AI-feedback methods point to a promising path for future self-adaptive systems, especially in learning user-aligned policies, reducing reliance on reward engineering through automated preference annotation.

\subsection*{Acknowledgements}
This publication has in part emanated from research supported by Taighde Eireann – Research Ireland under Frontiers for the Future Grant No. 21/FFP-A/8957. For the purpose of Open Access, the author has applied a CC BY public copyright license to any Author Accepted Manuscript version arising from this submission.

%%
%% The next two lines define the bibliography style to be used, and
%% the bibliography file.

% For the purpose of Open Access, the authors have applied a CC BY public copyright license to any Author Accepted Manuscript version arising from this submission.
\bibliographystyle{ACM-Reference-Format}
\bibliography{samples/seams26}

%%
%% If your work has an appendix, this is the place to put it.
% \appendix

\end{document}